\documentclass[runningheads]{llncs}
\usepackage[T1]{fontenc}
\usepackage{graphicx,verbatim}
\usepackage{color}
\usepackage{amsmath}
\usepackage{multirow}
\usepackage{amsfonts}       
\usepackage{cleveref}
\usepackage{makecell}
\usepackage{float}

\usepackage{marvosym}

\begin{document}
\title{Seeking Necessary and Sufficient Information from Multimodal Medical Data}
%

\author{
Boyu Chen\inst{1}\textsuperscript{*} \and
Weiye Bao\inst{2}\textsuperscript{*} \and
Junjie Liu\inst{2} \and
Michael Shen\inst{3} \and
Bo Peng\inst{1} \and
Paul Taylor\inst{1} \and
Zhu Li\inst{3}\textsuperscript{\Letter} \and
Mengyue Yang\inst{4}\textsuperscript{\Letter}
}

\authorrunning{B. Chen et al.}

\institute{
University College London, London, UK \\
\and
Imperial College London, London, UK \\
\and
Mingdu Tech, China \\
\and
University of Bristol, Bristol, UK \\
}
  
\maketitle              
%
\begingroup
\renewcommand{\thefootnote}{} 
\footnotetext{\textsuperscript{*} These authors contributed equally.}
\footnotetext{\textsuperscript{\Letter} Corresponding author.}
\endgroup

\begin{abstract}
Learning multimodal representations from medical images and other data sources can provide richer information for decision-making. While various multimodal models have been developed for this, they overlook learning features that are both necessary (must be present for the outcome to occur) and sufficient (enough to determine the outcome). We argue learning such features is crucial as they can improve model performance by capturing essential predictive information, and enhance model robustness to missing modalities as each modality can provide adequate predictive signals. Such features can be learned by leveraging the Probability of Necessity and Sufficiency (PNS) as a learning objective, an approach that has proven effective in unimodal settings. However, extending PNS to multimodal scenarios remains underexplored and is non-trivial as key conditions of PNS estimation are violated. We address this by decomposing multimodal representations into modality-invariant and modality-specific components, then deriving tractable PNS objectives for each. Experiments on synthetic and real-world medical datasets demonstrate our method's effectiveness. Code will be available on GitHub.

\keywords{Representation Learning \and Multi-modality \and Probability of Necessity and Sufficiency.}

\end{abstract}

\section{Introduction}
Multimodal representation learning has become essential in medical imaging, as integrating multiple data sources provides richer information for prediction. Various multimodal models have been developed for this task, such as multimodal fusion \cite{liu2023m3ae,zhang2022mmformer,ding2021rfnet,li2025multimodal,ma2025longitudinal}, contrastive learning \cite{lin2023pmc,zhang2023multi,wang2022medclip,zhang2025multimodal,gu2025learning}, and disentanglement \cite{chen2019robust,wang2023multi,li2025dc,eijpe2025disentangled}. However, they overlook the importance of learning the features that are both necessary and sufficient for the outcome.

Necessity means a feature must be present for the outcome to occur, but its presence alone does not confirm the outcome (e.g., pulmonary infiltrates on chest CT are typical in pneumonia, but infiltrates alone are not diagnostic).
Sufficiency means a feature can confirm the outcome when present, but it may be absent even if the outcome occurs (e.g., a visible pneumothorax line on chest X-ray confirms pneumothorax, but early pneumothorax may lack this line).
Ideally, predictive features are both necessary and sufficient (e.g., a fracture line on X-ray indicates a fracture, and fractures present with such lines).

Learning such features provides advantages for multimodal medical data. First, it improves predictive performance: necessity ensures critical information is captured, while sufficiency guarantees the captured information is predictive \cite{yang2023invariant,cai2023learning,chen2025medical}. Second, it can handle missing modalities, which is a pervasive challenge in clinical practice \cite{liu2023m3ae,zhang2022mmformer,ding2021rfnet,d2022fusing,li2025dc}. When each modality learns to capture necessary and sufficient features, any available subset of modalities retains adequate predictive signal, enabling robust inference with incomplete data.

Recent work has applied the Probability of Necessity and Sufficiency (PNS) to learn such features in unimodal data \cite{yang2023invariant,wang2024desiderata,chen2024unifying,cai2023learning,chen2025medical}. However, extending PNS to multimodal settings remains underexplored and is non-trivial as key conditions of PNS estimation are violated (details in \cref{sec:Preliminaries}). We address this by decomposing multimodal representations into modality-invariant and modality-specific components, then deriving tractable PNS objectives for each. Experiments on synthetic and real-world medical datasets show the effectiveness of our method. 


\section{Preliminaries}
\label{sec:Preliminaries}
This section covers PNS computation, challenges in multimodal extension, and representation decoupling—foundations for our method in \cref{sec:method}.

PNS is the probability of a feature set being both necessary and sufficient for an outcome. Let $Z\subset \mathbb{R}^{d_z}$ be the features of outcome $Y \subset \mathbb{R}^{d_y}$, where $z$ and $\bar{z}$ are two distinct values of $Z$. The PNS of $Z$ with respect to $Y$ for $z$ and $\bar{z}$ is defined as: $\text{PNS}(z, \bar{z}) := P(Y_{\mathrm{do}(Z=z)}=y|Z=\bar{z},Y \neq y)\cdot P(Z=\bar{z},Y \neq y) + P(Y_{\mathrm{do}(Z=\bar{z})}\neq y|Z=z,Y=y)\cdot P(Z=z,Y=y)$ \cite{pearl2009causality,yang2023invariant}.

The do-operator $\mathrm{do}(Z=z)$ denotes an intervention setting Z to z, which cannot be directly computed from observational data. However, under two conditions, PNS can be estimated using observational data:
\begin{definition}[Exogeneity \cite{pearl2009causality}]
$Z$ is exogenous to $Y$ if $Z$ has a direct effect on $Y$ without confounding factors, i.e., $P(Y_{do(Z=z)} = y) = P(Y = y \mid Z=z)$.
\end{definition}

\begin{definition}[Monotonicity \cite{pearl2009causality}]
$Y$ is monotonic w.r.t. $Z$ when $(Y_{do(Z=z)} \neq y) \wedge (Y_{do(Z=\bar{z})} = y)$ or $(Y_{do(Z=z)} = y) \wedge (Y_{do(Z=\bar{z})} \neq y)$ is false, which can be presented as: $P(Y_{do(Z=z)} = y)P(Y_{do(Z=\bar{z})} \neq y) = 0$ or $P(Y_{do(Z=z)} \neq y)P(Y_{do(Z=\bar{z})} = y) = 0$.
\label{def:mono}
\end{definition}

\begin{lemma}[\cite{pearl2009causality}]
\label{lemma:pns}
Under monotonicity and exogeneity:
\begin{equation} \nonumber
\begin{aligned}
\text{PNS}(z, \bar{z}) = & P(Y = y \mid Z = z) - P(Y = y \mid Z = \bar{z})\\
\end{aligned}
\end{equation}
\end{lemma}

Recent studies have applied PNS to guide representation learning in unimodal data \cite{yang2023invariant,wang2024desiderata,chen2024unifying,cai2023learning,chen2025medical}, but extending it to multimodal scenarios faces challenges in satisfying these two conditions. Exogeneity is violated since interactions across modalities can introduce hidden confounding effects without strong assumptions \cite{yang2023invariant,wang2024desiderata,locatello2019challenging,liu2021learning}.
Without exogeneity, the interventional probabilities in \cref{def:mono} cannot be converted to conditional probabilities, making it intractable to establish monotonicity from observational data.

\begin{figure*}[ht]
\centerline{\includegraphics[width=1\textwidth,trim=55 85 50 85,clip]{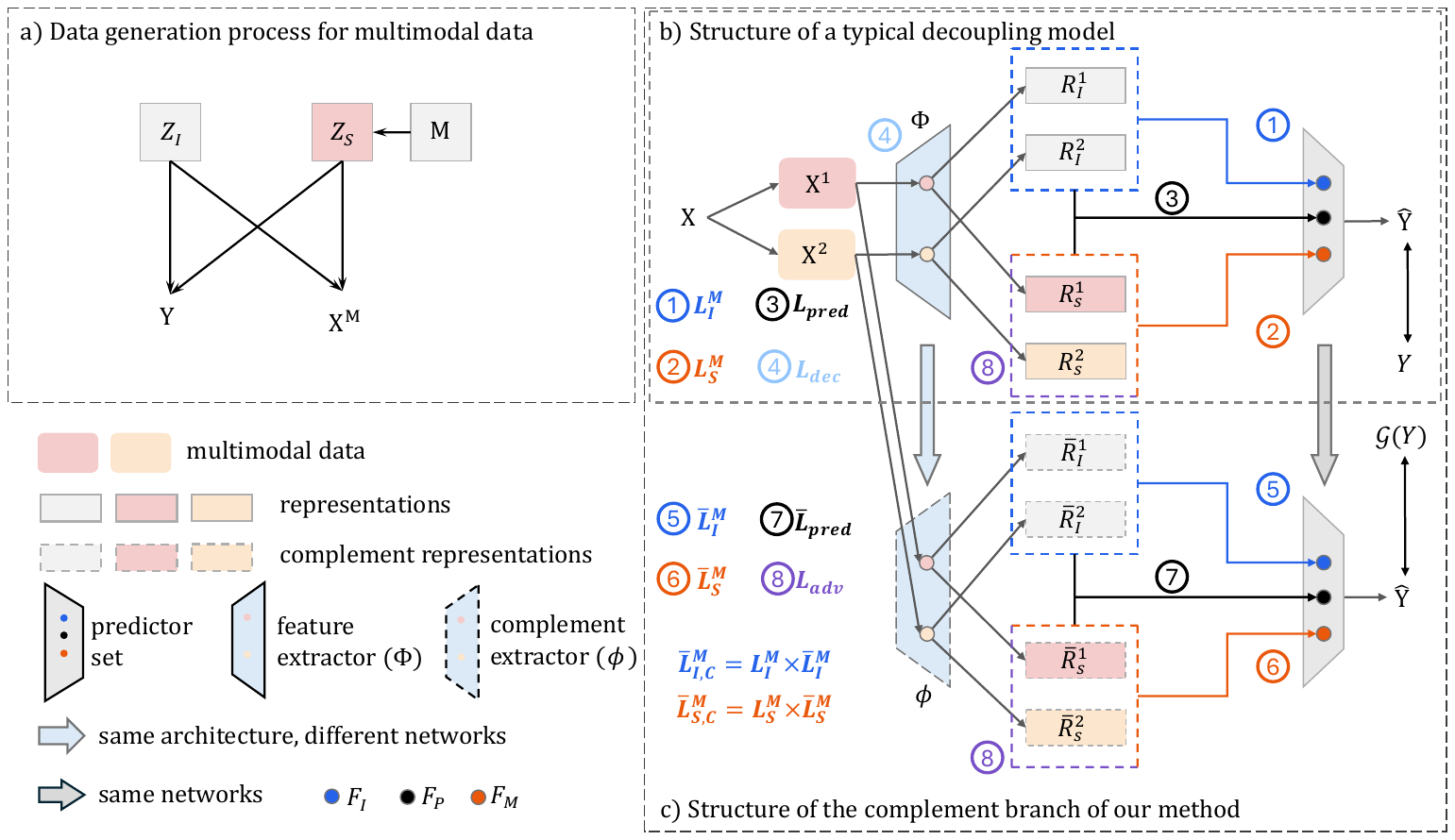}}
\caption{Overview of the proposed MPNS framework. The figure shows the multimodal data generation process, a baseline decoupling model, and our proposed complement branch with PNS-based optimization.}
\label{fig:method}
\end{figure*}

To address these challenges, our work builds upon the assumption that multimodal data can be decoupled into two types of latent variables: one shared across modalities and the other conditioned on the modality type \cite{chen2019robust,wang2023multi,li2025dc,li2023decoupled}. As illustrated in \cref{fig:method}(a), the data generation process involves two latent factors. Let $(X^1,...,X^N,Y)$ be a multimodal sample with $N$ modalities, abbreviated as $(X,Y)$, where $X^M \subset \mathbb{R}^{d_M}$ denotes the input from modality $M$ and $Y \subset \mathbb{R}^{d_y}$ denotes the outcome. A modality sample ($X^M$, $Y$) is generated from: a modality-invariant latent variable $Z_I \subset \mathbb{R}^{d_{Z_I}}$ capturing shared information across modalities, and modality-specific latent variables $Z_S \subset \mathbb{R}^{d_{Z_S}}$ conditioned on modality type $M$, encoding characteristics unique to each modality \cite{chen2019robust,wang2023multi,li2025dc,li2023decoupled}.

A decoupling (or disentanglement) $\mathcal{D}(\cdot)$ model aims to capture information from these latent factors by 
extracting two types of representations from $X^M$: modality-invariant representations 
$\mathcal{R}_I^M \subset \mathbb{R}^{d_{Z_I}}$ approximating $Z_I$, and 
modality-specific representations $\mathcal{R}_S^M \subset \mathbb{R}^{d_{Z_S}}$ 
approximating $Z_S$ conditioned on modality type $M$. 
A typical architecture of $\mathcal{D}(\cdot)$ (shown in \cref{fig:method}(b)) which consists of a feature extractor $\Phi(\cdot)$ that decouples the representations $[\mathcal{R}_{I}^M, \mathcal{R}_{S}^M]:= \phi(X^M)$, and three types of predictors: (1) modality-specific predictors $F_{M}(\cdot)$ predicting $Y$ from $\mathcal{R}_S^M$; (2) an invariant predictor $F_{I}(\cdot)$ predicting $Y$ from $\mathcal{R}_I^M$; (3) a joint predictor $F_{P}(\cdot)$ predicting $Y$ from all representations. We denote the overall model prediction as $\hat{Y} = \mathcal{D}(X)$. The training objective is:

\begin{equation}
\label{eq:decouple-loss}
\mathcal{L}^{D} := \mathcal{L}_{pred} + \mathcal{L}_{dec} + \textstyle\sum_{M=1}^{N}\left( \mathcal{L}_{I}^{M} + \mathcal{L}_{S}^{M}\right)
\end{equation}
where $\mathcal{L}_{I}^{M} := \mathcal{L}_{I}^{M}(F_{I}(\mathcal{R}_{I}^M), Y)$ and $\mathcal{L}_{S}^{M} := \mathcal{L}_{S}^{M}(F_{M}(\mathcal{R}_{S}^M), Y)$ measure the prediction accuracy of $\mathcal{R}_{I}^M$ and $\mathcal{R}_{S}^M$, respectively, while $\mathcal{L}_{pred} := \mathcal{L}_{pred}(\mathcal{D}(X), Y)$ is the prediction loss, and $\mathcal{L}_{dec}$ includes decoupling and other constraints. The specific architectures of feature extractors, predictors, and loss functions vary across prior multimodal decoupling work.

\section{Method}
\label{sec:method}
Following \cref{sec:Preliminaries}, we decompose representations into invariant and specific components, enabling separate PNS analysis for each.

\subsection{Learning Complement Representations}
Computing PNS via \cref{lemma:pns} requires feature pairs $(z, \bar{z})$ where $z$ predicts correctly while $\bar{z}$ does not. We achieve this by generating complement representations $\bar{\mathcal{R}}_{I}^M$ and $\bar{\mathcal{R}}_{S}^M$ that predict incorrectly, in contrast to $\mathcal{R}_{I}^M$ and $\mathcal{R}_{S}^M$. Specifically, as shown in \cref{fig:method}(c), we use a complement feature extractor $\phi(\cdot)$ that mirrors the architecture of $\Phi(\cdot)$ but extracts $[\bar{\mathcal{R}}_{I}^M, \bar{\mathcal{R}}_{S}^M]:= \phi(X^M)$. To train $\phi(\cdot)$ for incorrect predictions, we define a label generator $\mathcal{G}(\cdot)$ that randomly samples labels different from $Y$: $\bar{Y} = \mathcal{G}(Y)$. The complement extractor is optimized via: 
\begin{equation}
\bar{\mathcal{L}}_{pred} := \mathcal{L}_{pred}(\mathcal{D}_{\phi}(X), \mathcal{G}(Y))
\end{equation}
where $\mathcal{D}_{\phi}(\cdot)$ denotes the decoupling model using $\phi(\cdot)$.

\subsection{PNS for Modality-Invariant Representation}
As illustrated in \cref{fig:method}(a), analyzing $Z_I$ alone eliminates cross-modal confounding, making it exogenous to $Y$ since no other variable acts as a confounder between them. Under monotonicity, \cref{lemma:pns} enables tractable PNS estimation from observational data. This reduces to the unimodal setting, and therefore we adopt the PNS-based objective 
from \cite{yang2023invariant} for the invariant component of modality $M$, which can capture necessary and sufficient information:

\begin{equation}
\label{eq:pns_I_loss}
\mathcal{L}_{M,I}^{pns} :=  \mathcal{L}_{I}^{M} + \bar{\mathcal{L}}_{I}^{M} + \mathcal{L}_{I,C}^{M}
\end{equation}
where $\mathcal{L}_{I}^{M}$ (from \cref{eq:decouple-loss}) maximizes $P(Y=y \mid Z_I=z)$ by training $F_I(\mathcal{R}_I^M)$ to predict $Y$ accurately, and $\bar{\mathcal{L}}_{I}^{M} := \mathcal{L}_{I}^{M}(F_I(\bar{\mathcal{R}}_{I}^M),\mathcal{G}(Y))$ minimizes $P(Y=y \mid Z_I=\bar{z})$ by enforcing $F_I(\bar{\mathcal{R}}_I^M)$ 
predicts away from $Y$. 

The constraint term $\mathcal{L}_{I,C}^{M} := \mathcal{L}_{I}^{M} \times \bar{\mathcal{L}}_{I}^{M}$ enforces monotonicity by minimizing $P(Y \neq y|Z_I=z) \cdot P(Y = y|Z_I=\bar{z}) = P(Y_{do(Z_I=z)} \neq y) \cdot P(Y_{do(Z_I=\bar{z})} = y)$ from \cref{def:mono}.

\subsection{PNS for Modality-Specific Representation}
Unlike the modality-invariant component, the modality-specific component is conditioned on M, violating exogeneity (\cref{fig:method}(a)). As such, we propose to approximate exogeneity by enforcing $\mathcal{R}_S^M \perp M$ on the learned representations.

We use adversarial training to achieve this. Specifically, a modality discriminator is trained to classify $M$ from $\mathcal{R}_S^M$, while feature extractors $\Phi(\cdot)$ and $\phi(\cdot)$ simultaneously learn to produce representations that fool the discriminator via gradient reversal layer (GRL) \cite{ganin2016domain}, and maintaining predictive power for outcomes. We define the adversarial loss as $\mathcal{L}_{adv}$.

Note that $\mathcal{R}_S^M \perp M$ removes dependence on modality identity (preventing the model from learning "this is modality $M$") but not on modality-specific content. Unlike $\mathcal{R}_I$, which captures shared information across modalities, $\mathcal{R}_S^M$ still encodes information unique to that modality via $\mathcal{D}(\cdot)$. The independence constraint encourages $\mathcal{R}_S^M$ to learn these unique features based on their diagnostic value rather than on artifacts of modality type $M$.

When we break the dependency between $\mathcal{R}_S^M$ and $M$, \cref{lemma:pns} can be applied analogously to the modality-invariant case. The PNS objective for $M$ is:

\begin{equation}
\label{eq:pns_S_loss}
\mathcal{L}_{M,S}^{pns} :=  \mathcal{L}_{S}^{M} + \bar{\mathcal{L}}_{S}^{M} + \mathcal{L}_{S,C}^{M}
\end{equation}
where $\mathcal{L}_{S}^{M}$ (from \cref{eq:decouple-loss}) encourages high $P(Y=y \mid Z_S=z)$ by 
training $F_M(\mathcal{R}_S^M)$ to predict $Y$ accurately, $\bar{\mathcal{L}}_{S}^{M} := \mathcal{L}_{S}^{M}(F_M(\bar{\mathcal{R}}_S^M), \mathcal{G}(Y))$ minimizes $P(Y=y \mid Z_S=\bar{z})$, and $\mathcal{L}_{S,C}^{M} := \mathcal{L}_{S}^{M} \times \bar{\mathcal{L}}_{S}^{M}$ enforces monotonicity, mirroring \cref{eq:pns_I_loss}. Some decoupling 
models may not include modality-specific $F_M(\cdot)$. In such cases, $F_M(\cdot)$ can be implemented by adapting the architecture of $F_I(\cdot)$ or $F_P(\cdot)$.

\subsection{Multimodal Representation Learning via PNS}
Our framework are shown in \cref{fig:method}(c) and the final objective is:
\begin{equation}
\label{eq:pns_loss}
\mathcal{L}^{pns} := \mathcal{L}^{D} + \bar{\mathcal{L}}_{pred} + \mathcal{L}_{adv} + \textstyle\sum_{M=1}^{N}\left( \bar{\mathcal{L}}_{I}^{M} + \mathcal{L}_{I,C}^{M} + \bar{\mathcal{L}}_{S}^{M} + \mathcal{L}_{S,C}^{M}\right)
\end{equation}

We name our approach MPNS (\underline{M}ultimodal Representation Learning via \underline{PNS}). The implementation is straightforward (\cref{fig:method}(c)): add a complement extractor $\phi(\cdot)$ and the adversarial training to the base model $\mathcal{D}(\cdot)$, and optimize $\mathcal{L}^{pns}$ jointly. At inference, only $\mathcal{D}(\cdot)$ is retained. This makes MPNS a plug-and-play framework that introduces no additional inference cost.

\section{Experiments and Results}
\label{sec:experiment}
We constructed synthetic datasets to verify the ability of MPNS to capture necessary and sufficient information for representations. Then, we used real-world multimodal MRI datasets to show the improvement and robustness brought by MPNS. All experiments were conducted on 4 $\times$ NVIDIA GH200 GPUs.

\begin{table*}[t]
\caption{Distance Correlation based on $s$ for modality 1 and 2}
\label{tab:syn-dataset-results-1}
\centering
\setlength{\tabcolsep}{1.5pt}
\renewcommand{\arraystretch}{0.9}
\begin{tabular}{ll|cccc|cccc}
\hline
& \multirow{2}{*}{Mode} & \multicolumn{4}{c|}{Modality 1} & \multicolumn{4}{c}{Modality 2} \\
\cline{3-10}
& & NS & SF & NC & SC & NS & SF & NC & SC \\
\hline
\multirow{4}{*}{s = 0.0}
& w/o PNS & 0.626 & 0.667 & 0.658 & 0.291 & 0.519 & 0.607 & 0.646 & 0.312 \\
& w/o  $\mathcal{L}_{M,I}^{pns}$ & 0.645 & 0.681 & 0.665 & 0.294 & 0.536 & 0.597 & 0.633 & 0.323 \\
& w/o  $\mathcal{L}_{M,S}^{pns}$ & 0.663 & 0.673 & 0.644 & 0.305 & 0.627 & 0.577& 0.636 & 0.349\\
& MPNS & \textbf{0.682} & 0.658 & 0.578 & 0.298 & \textbf{0.649} & 0.553 & 0.629 & 0.364 \\
\hline
\multirow{4}{*}{s = 0.3}
& w/o PNS & 0.613 & 0.682 & 0.645 & 0.322 & 0.520 & 0.618 & 0.614 & 0.347 \\
& w/o  $\mathcal{L}_{M,I}^{pns}$ & 0.631 & 0.683 & 0.643 & 0.316 & 0.541 & 0.607 & 0.608 & 0.343 \\
& w/o  $\mathcal{L}_{M,S}^{pns}$ & 0.644 & 0.691 & 0.591& 0.322 & 0.621 & 0.591& 0.617 & 0.377\\
& MPNS & \textbf{0.661} & 0.657 & 0.572 & 0.323 & \textbf{0.640} & 0.564 & 0.617 & 0.394 \\
\hline
\multirow{4}{*}{s = 0.7}
& w/o PNS & 0.619 & 0.683 & 0.664 & 0.352 & 0.503 & 0.614 & 0.636 & 0.371 \\
& w/o  $\mathcal{L}_{M,I}^{pns}$ & 0.635 & 0.696 & 0.678 & 0.361 & 0.521 & 0.602 & 0.642 & 0.369 \\
& w/o  $\mathcal{L}_{M,S}^{pns}$ & 0.666 & 0.683 & 0.602& 0.364 & 0.633 & 0.574& 0.630 & 0.412\\
& MPNS & \textbf{0.678} & 0.652 & 0.585 & 0.361 & \textbf{0.647} & 0.548 & 0.599 & 0.448 \\
\hline
\end{tabular}
\end{table*}

\subsection{Synthetic Multimodal Dataset Experiments}
To verify MPNS's ability to learn high-PNS representations, we followed the protocol from \cite{yang2023invariant}: 1) generate a variable, called necessary and sufficient variable $NS$, that directly determines the outcome; 2) generate other variables with different properties; 3) mix them as latent variables; 4) transform them by nonlinear functions to create observed data; 5) train models on the observed data to learn representations; 6) measure how well representations capture the original latent variables via Distance Correlation (DC) \cite{jones1995fitness}, where higher DC with $NS$ indicates better high-PNS learning. We extend this to multimodal settings:

We first generated the latent variables and outcome using the same method as \cite{yang2023invariant}:
\textbf{Necessary and Sufficient variable $NS$} is a deterministic variable sampled from a Bernoulli distribution $B(0.5)$ and determines the outcome $Y=NS\oplus B(0.15)$, where $\oplus$ is the XOR operation.
\textbf{Sufficient but Unnecessary variable} $SF \sim B(0.1)$ if $NS$=0; $SF$=1 if $NS$=1.
\textbf{Necessary but Insufficient variable} $NC =  I(NS=1)\cdot B(0.9)$, where $I(\cdot)$ is an indicator function. \textbf{Spurious correlation variable $SC$} is generated to spuriously correlate with $NS$, defined as $s \cdot NS + (1 - s) \mathcal{N}(0, 1)$, where $s \in [0,1)$ is spurious correlation level and $\mathcal{N}(0, 1)$ is a standard Gaussian distribution. 

Then, We generated observed data with two modalities from latent variables. We construct a vector $h = [NS\cdot \mathbf{1}_d, SF\cdot \mathbf{1}_d, NC\cdot \mathbf{1}_d, SC\cdot \mathbf{1}_d] + \mathcal{N}(0,0.3)$, where $\mathbf{1}_d$ is a $d$-dimensional all-ones vector and $Y$ is determined by $NS$. For each variable type, we use the first, middle, and last $d/3$ dimensions of each type as modality-invariant ($z_1$) and modality-specific ($z_2$ and $z_3$) components. The nonlinear function is $\kappa(z, \beta) = \beta \cdot \tanh(z)$ , and $[\cdot,\cdot]$ denotes concatenation of column vectors. The two modalities are then generated as $X^1 = \kappa([z_1, \kappa(z_2, \beta_1)], \beta_2)$ and $X^2 = \kappa([z_1, \kappa(z_3, \beta_3)], \beta_4)$. We set $(\beta_1, \beta_2, \beta_3, \beta_4) = (2.0, 1.8, 1.5, 1.2)$ and $d=15$.

Next, we trained models on the observed data. We used DMD \cite{li2023decoupled}, a state-of-the-art (SOTA) multimodal decoupling model, as the base model of MPNS. We also conducted ablation studies: 1) \textbf{w/o PNS}. original DMD is optimized only with $\mathcal{L}^{D}$ in \cref{eq:decouple-loss}; 2) \textbf{w/o $\mathcal{L}_{M,I}^{pns}$}. only $\mathcal{L}_{I}^{M}$ is optimized in \cref{eq:pns_I_loss}; 3) \textbf{w/o $\mathcal{L}_{M,S}^{pns}$}. only $\mathcal{L}_{S}^{M}$ is optimized in \cref{eq:pns_S_loss}; and 4) \textbf{MPNS}. the full model trained with complete $\mathcal{L}^{pns}$. We varied $s$ as 0.0, 0.3 and 0.7 to show the different spurious correlation levels. We generated 15,000 training and 5,000 evaluation samples for each $s$. Results are shown on \cref{tab:syn-dataset-results-1}.

\begin{table*}[!t]
\centering
\caption{Dice coefficients on BraTS2020 with missing modalities. Notes: $\bullet$/$\circ$: available/missing modality. \textbf{Bold}/\underline{underlined}: \textbf{best}/\underline{second-best} result. $\dagger$: MPNS outperforms its base decoupling model.}
\label{tab:complete_dice_result}
\setlength{\tabcolsep}{1.5pt}
\renewcommand{\arraystretch}{0.9}
\resizebox{\textwidth}{!}{
\begin{tabular}{c|c c c c|c c c c c c c c}
\hline
\textbf{\thead{Subregions}} &
\textbf{\thead{F}} & \textbf{\thead{T1}} & \textbf{\thead{T1c}} & \textbf{\thead{T2}} &
\textbf{\thead{RobustSeg}} & \textbf{\thead{RFNet}} & \textbf{\thead{mmFormer}} & \textbf{\thead{M$^3$AE}} &
\textbf{\thead{ShaSpec}} &
 \textbf{\thead{ShaSpec\\(MPNS)}} 
&\textbf{\thead{DC-Seg}} 
& \textbf{\thead{DC-Seg\\(MPNS)}} \\
\hline

\multirow{16}{*}{\textbf{Whole}} & $\circ$ & $\circ$ & $\circ$ & $\bullet$ & 82.20 & \underline{86.05}& 85.51 & \textbf{86.10}& 83.98 &  83.86 &85.16 & 85.09\\
& $\circ$ & $\circ$ & $\bullet$ & $\circ$ & 71.39 & 76.77 & 78.04 & \underline{78.90}& 76.18 &  76.91$^{\dagger}$& 78.57 & \textbf{79.45}$^{\dagger}$\\
& $\circ$ & $\bullet$ & $\circ$ & $\circ$ & 71.41 & 77.16 & 76.24 & \textbf{79.00}& 74.27 &  74.65$^{\dagger}$& \underline{77.33}& 76.96 \\
& $\bullet$ & $\circ$ & $\circ$ & $\circ$ & 82.87 & 87.32 & 86.54 & 88.00 & \textbf{89.25}&  \textbf{89.25}& 88.86 & \underline{89.01}$^{\dagger}$\\
& $\circ$ & $\circ$ & $\bullet$ & $\bullet$ & 85.97 & 87.74 & 87.52 & 87.10 & 86.57 &  86.21 & \underline{87.94}& \textbf{88.07}$^{\dagger}$\\
& $\circ$ & $\bullet$ & $\bullet$ & $\circ$ & 76.84 & 81.12 & 80.70 & 80.10 & 78.92 &  78.90 & \underline{81.99}& \textbf{82.00}$^{\dagger}$\\
& $\bullet$ & $\bullet$ & $\circ$ & $\circ$ & 88.10 & 89.73 & 88.76 & 89.60 & \textbf{90.31}&  \textbf{90.31}& 90.03 & \underline{90.06}$^{\dagger}$\\
& $\circ$ & $\bullet$ & $\circ$ & $\bullet$ & 85.53 & \textbf{87.73}& 86.94 & 87.30 & 85.42 &  85.77$^{\dagger}$& 87.13 & \underline{87.39}$^{\dagger}$\\
& $\bullet$ & $\circ$ & $\circ$ & $\bullet$ & 88.09 & 89.87 & 89.49 & 90.10 & \underline{90.43}&  90.41 & 90.06 & \textbf{91.07}$^{\dagger}$\\
& $\bullet$ & $\circ$ & $\bullet$ & $\circ$ & 87.33 & 89.89 & 89.31 & 89.50 & 90.34 &  \textbf{90.43}$^{\dagger}$& 90.31 & \underline{90.36}$^{\dagger}$\\
& $\bullet$ & $\bullet$ & $\bullet$ & $\circ$ & 88.87 & 90.69 & 89.79 & 89.60 & 90.60 &  90.54 & \underline{90.83}& \textbf{90.90}$^{\dagger}$\\
& $\bullet$ & $\bullet$ & $\circ$ & $\bullet$ & 89.24 & 90.60 & 89.83 & 90.20 & \underline{90.65}&  90.61 & 90.59 & \textbf{90.74}$^{\dagger}$\\
& $\bullet$ & $\circ$ & $\bullet$ & $\bullet$ & 88.68 & 90.68 & 90.49 & 90.50 & 90.89 &  \underline{90.90}$^{\dagger}$& 90.76 & \textbf{91.35}$^{\dagger}$\\
& $\circ$ & $\bullet$ & $\bullet$ & $\bullet$ & 86.63 & \textbf{88.25}& 87.64 & 87.40 & 86.94 &  86.49 & \underline{88.24}& 88.23 \\
& $\bullet$ & $\bullet$ & $\bullet$ & $\bullet$ & 89.47 & \textbf{91.11}& 90.54 & 90.40 & \underline{91.04}&  90.91 & 90.93 & 90.94$^{\dagger}$\\
\cline{2-13}
& \multicolumn{4}{c|}{\textbf{Average}} & 84.17 & 86.98 & 86.49 & 86.90 & 86.39 &  86.41$^{\dagger}$& \underline{87.25} & \textbf{87.44}$^{\dagger}$\\

\hline

\multirow{16}{*}{\textbf{Core}} & $\circ$ & $\circ$ & $\circ$ & $\bullet$ & 61.88 & \underline{71.02}& 63.36 & \textbf{71.80}& 65.57 &  65.58$^{\dagger}$&68.09 & 69.58$^{\dagger}$\\
& $\circ$ & $\circ$ & $\bullet$ & $\circ$ & 76.68 & 81.51 & 81.51 & \textbf{83.60}& 80.60 &  80.87$^{\dagger}$& \underline{83.31}& 83.30\\
& $\circ$ & $\bullet$ & $\circ$ & $\circ$ & 54.30 & 66.02 & 63.23 & \textbf{69.40}& 63.72 &  65.20$^{\dagger}$& 65.85 & \underline{66.14}$^{\dagger}$\\
& $\bullet$ & $\circ$ & $\circ$ & $\circ$ & 60.72 & 69.19 & 64.60 & 68.70 & 69.15 &  69.79$^{\dagger}$& \underline{70.26}& \textbf{71.40}$^{\dagger}$\\
& $\circ$ & $\circ$ & $\bullet$ & $\bullet$ & 82.44 & 83.45 & 82.69 & \textbf{85.60}& 81.94 &  81.66 & 83.82 & \underline{84.38}$^{\dagger}$\\
& $\circ$ & $\bullet$ & $\bullet$ & $\circ$ & 80.28 & 83.40 & 82.81 & 83.80 & 82.01 &  81.45 & \underline{84.35}& \textbf{84.40}$^{\dagger}$\\
& $\bullet$ & $\bullet$ & $\circ$ & $\circ$ & 68.18 & 73.07 & 71.76 & 72.80 & 72.93 &  \underline{73.37}$^{\dagger}$& \textbf{73.81}& 72.98 \\
& $\circ$ & $\bullet$ & $\circ$ & $\bullet$ & 66.46 & \textbf{73.13}& 67.76 & \underline{72.90}& 68.51 &  69.36$^{\dagger}$& 70.40 & 71.70$^{\dagger}$\\
& $\bullet$ & $\circ$ & $\circ$ & $\bullet$ & 68.20 & 74.14 & 70.34 & \underline{74.30}& 71.58 &  71.97$^{\dagger}$& 73.43 & \textbf{75.33}$^{\dagger}$\\
& $\bullet$ & $\circ$ & $\bullet$ & $\circ$ & 81.85 & \underline{84.65}& 83.79 & \textbf{85.50}& 82.39 &  82.37 & \underline{84.65}& 83.91 \\
& $\bullet$ & $\bullet$ & $\bullet$ & $\circ$ & 82.76 & \underline{85.07}& 84.44 & \textbf{85.60}& 83.08 &  82.83 & 84.88 & 83.84 \\
& $\bullet$ & $\bullet$ & $\circ$ & $\bullet$ & 70.46 & \underline{75.19}& 72.42 & 74.40 & 72.54 &  73.13$^{\dagger}$& 73.75 & \textbf{75.47}$^{\dagger}$\\
& $\bullet$ & $\circ$ & $\bullet$ & $\bullet$ & 81.89 & \underline{84.97}& 83.94 & \textbf{85.80}& 82.44 &  82.00 & 84.25 & 84.63$^{\dagger}$\\
& $\circ$ & $\bullet$ & $\bullet$ & $\bullet$ & 82.85 & 83.47 & 83.66 & \textbf{85.80}& 82.70 &  82.07 & \underline{84.70}& 84.37\\
& $\bullet$ & $\bullet$ & $\bullet$ & $\bullet$ & 82.87 & \underline{85.21}& 84.61 & \textbf{86.20}& 82.98 &  82.38 & 84.27 & 84.49$^{\dagger}$\\
\cline{2-13}
& \multicolumn{4}{c|}{\textbf{Average}} & 73.45 & 78.23 & 76.06 & \textbf{79.10} & 76.14 &  76.27$^{\dagger}$& 77.99 & \underline{78.39}$^{\dagger}$\\

\hline

\multirow{16}{*}{\textbf{Enhancing}} & $\circ$ & $\circ$ & $\circ$ & $\bullet$ & 36.46 & 46.29 & \textbf{49.09}& 47.10 & 42.31 &  41.22 &47.14 & \underline{47.35}$^{\dagger}$\\
& $\circ$ & $\circ$ & $\bullet$ & $\circ$ & 67.91 & 74.85 & \textbf{78.30}& 73.60 & 72.83 &  72.87$^{\dagger}$& 76.68 & \underline{76.74}$^{\dagger}$\\
& $\circ$ & $\bullet$ & $\circ$ & $\circ$ & 28.99 & 37.30 & 37.62 & 40.40 & 38.61 &  39.52$^{\dagger}$& \underline{40.73}& \textbf{42.47}$^{\dagger}$\\
& $\bullet$ & $\circ$ & $\circ$ & $\circ$ & 34.68 & 38.15 & 36.68 & 40.20 & 44.09 &  43.54 & \textbf{44.98}& \underline{44.29}\\
& $\circ$ & $\circ$ & $\bullet$ & $\bullet$ & 71.42 & 75.93 & \textbf{77.20}& 76.00 & 73.95 &  73.56 & 76.09 & \underline{76.68}$^{\dagger}$\\
& $\circ$ & $\bullet$ & $\bullet$ & $\circ$ & 70.11 & \underline{78.01}& \textbf{81.71}& 75.30 & 73.40 &  73.55$^{\dagger}$& 76.96 & 77.25$^{\dagger}$\\
& $\bullet$ & $\bullet$ & $\circ$ & $\circ$ & 39.67 & 40.98 & 42.98 & 43.70 & 46.65 &  47.54$^{\dagger}$& \underline{47.97}& \textbf{49.57}$^{\dagger}$\\
& $\circ$ & $\bullet$ & $\circ$ & $\bullet$ & 39.92 & 45.65 & \underline{49.12}& 48.70 & 45.28 &  45.04 & 48.76 & \textbf{51.27}$^{\dagger}$\\
& $\bullet$ & $\circ$ & $\circ$ & $\bullet$ & 42.19 & 49.32 & 49.06 & 47.10 & 46.25 &  46.72$^{\dagger}$& \textbf{52.01}& \underline{51.98}\\
& $\bullet$ & $\circ$ & $\bullet$ & $\circ$ & 70.78 & 76.67 & \textbf{79.44}& 75.90 & 74.12 &  74.25$^{\dagger}$& 76.66 & \underline{77.01}$^{\dagger}$\\
& $\bullet$ & $\bullet$ & $\bullet$ & $\circ$ & 71.77 & 76.81 & \textbf{80.65}& 76.30 & 74.17 & 74.59$^{\dagger}$& \underline{77.74}& 77.50\\
& $\bullet$ & $\bullet$ & $\circ$ & $\bullet$ & 43.90 & 49.92 & 50.08 & 48.20 & 47.62 & 48.69$^{\dagger}$& \underline{52.84}& \textbf{55.43}$^{\dagger}$\\
& $\bullet$ & $\circ$ & $\bullet$ & $\bullet$ & 71.17 & 77.12 & \textbf{78.73}& \underline{77.40}& 73.95 & 73.93 & 76.57 & 77.25$^{\dagger}$\\
& $\circ$ & $\bullet$ & $\bullet$ & $\bullet$ & 71.87 & \underline{76.99}& 77.34 & \textbf{78.00}& 74.13 & 73.96 & 76.16 & 76.28$^{\dagger}$\\
& $\bullet$ & $\bullet$ & $\bullet$ & $\bullet$ & 71.52 & \underline{78.00}& \textbf{79.92}& 77.50 & 73.96 & 74.37$^{\dagger}$& 76.83 & 76.43\\
\cline{2-13}
& \multicolumn{4}{c|}{\textbf{Average}} & 55.49 & 61.47 & 63.19 & 61.70 & 60.09 &  60.22$^{\dagger}$& \underline{63.21} & \textbf{63.83}$^{\dagger}$\\

\hline
\multicolumn{5}{c|}{\textbf{Average}} &
71.04 & 75.56 & 75.25 & 75.90 & 74.21 & 74.30$^{\dagger}$& \underline{76.15} & \textbf{76.55}$^{\dagger}$\\

\hline

\end{tabular}}
\end{table*}

\subsection{Real-world Multimodal Medical Dataset Experiments}
We evaluated MPNS on BraTS2020 \cite{zadeh2018multimodal}, a widely used multimodal MRI dataset for brain tumor segmentation. It contains four MRI modalities (FLAIR, T1c, T1, T2) with segmentation masks for three tumor sub-regions (whole tumor, tumor core, and enhancing tumor), and has 369 cases split into 219/50/100 for training/validation/testing. We compared against SOTA and open-sourced multimodal medical segmentation methods: RobustSeg \cite{chen2019robust}, RFNet \cite{ding2021rfnet}, mmFormer \cite{zhang2022mmformer}, M$^3$AE \cite{liu2023m3ae}, ShaSpec \cite{wang2023multi}, and DC-Seg \cite{li2025dc}. Among these, ShaSpec and DC-Seg are decoupling models, and serve as our base architectures of MPNS. Results are shown on Table~\ref{tab:complete_dice_result}.

\subsection{Results and Discussion}
Table \ref{tab:syn-dataset-results-1} shows DC between representations and latent variables, where higher DC with $NS$ indicates better representation. DMD with MPNS consistently outperforms all variants. As $s$ increases, DC with $SC$ also rises for all models but MPNS still has a strong correlation with $NS$. Removing $\mathcal{L}_{M,I}^{pns}$ causes more performance drop than removing $\mathcal{L}_{M,S}^{pns}$, suggesting that our method can capture more necessary and sufficient information from the modality-invariant component in data. The reason could be that invariant components naturally satisfy exogeneity while specific components need approximation. These results show that our MPNS can capture more necessary and sufficient information from data.

Table~\ref{tab:complete_dice_result} compares our method against baselines under missing modality scenarios. MPNS improves the performance of both base decoupling models (ShaSpec and DC-Seg), outperforming the baselines in most cases. This shows that learning high-PNS representations can enhance both the performance and robustness of multimodal models.

Our work introduces PNS to multimodal representation learning and could open promising avenues for future research, despite certain limitations. First, our method's effectiveness depends on the base decoupling model's disentanglement quality. When decoupling is suboptimal (e.g., for the core region in \cref{tab:complete_dice_result}), our performance gains are limited. As multimodal disentanglement remains an active research area, advances in decoupling models could further enhance our approach. Second, we approximate exogeneity for modality-specific representations via adversarial training, but alternative strategies merit exploration. Third, our method captures necessary and sufficient features within individual modalities, while such features may also emerge from cross-modal combinations. Fourth, our current formulation focuses on discrete outcomes, while extending to continuous outcomes requires further investigation. Extending PNS to capture these joint patterns represents a promising direction. By introducing PNS to multimodal learning, our work provides insights for future research in multimodal representation learning and related domains.

\section{Conclusion}
\label{sec:conclusion}
We introduce PNS to guide multimodal medical representation learning and propose a method that promotes high-PNS representations. Experiments on synthetic and real-world datasets show that our method can learn high-PNS representations and improve the performance and robustness of multimodal models.

\newpage
%
%
%
\bibliographystyle{splncs04}
\bibliography{mybibliography}
%




\end{document}